\documentclass{article}

\usepackage[final]{corl_2020} 

\usepackage{times}
\usepackage{epsfig}
\usepackage{graphicx}
\usepackage{amsmath}
\usepackage{amssymb}
\usepackage{caption}
\usepackage{tabularx}
\usepackage{multirow}

\title{One Thousand and One Hours:\\Self-driving Motion Prediction Dataset}

\author{
  John Houston \hspace{5mm} Guido Zuidhof \hspace{5mm} Luca Bergamini \hspace{5mm} Yawei Ye\\ \textbf{Long Chen \hspace{5mm} Ashesh Jain \hspace{5mm} Sammy Omari \hspace{5mm} Vladimir Iglovikov \hspace{5mm} Peter Ondruska}\\
  Lyft Level 5\\
  \texttt{level5data@lyft.com}
}

\begin{document}
\maketitle


\begin{figure}[h]
    \centering
    \includegraphics[width=\textwidth]{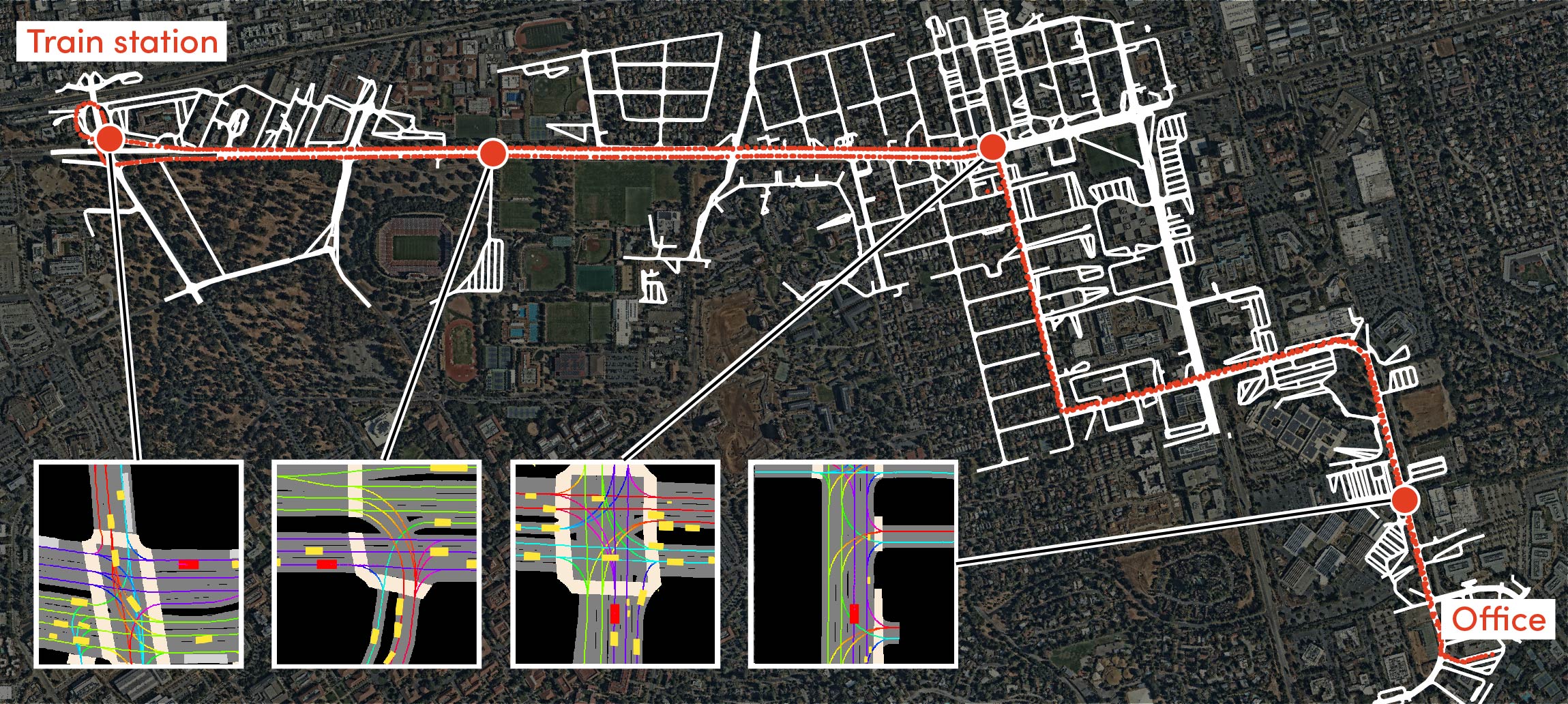}
    \captionof{figure}{An overview of the released dataset for motion modelling, consisting of 1,118 hours of recorded self-driving perception data on a route spanning 6.8 miles between the train station and the office (red). The examples on the bottom-left show released scenes on top of the high-definition semantic map that capture road geometries and the aerial view of the area.}
    \label{fig:intro}
\end{figure}

\begin{abstract}
Motivated by the impact of large-scale datasets on ML systems we present the largest self-driving dataset for motion prediction to date, containing over 1,000 hours of data. This was collected by a fleet of 20 autonomous vehicles along a fixed route in Palo Alto, California, over a four-month period. It consists of 170,000 scenes, where each scene is 25 seconds long and captures the perception output of the self-driving system, which encodes the precise positions and motions of nearby vehicles, cyclists, and pedestrians over time. On top of this, the dataset contains a high-definition semantic map with 15,242 labelled elements and a high-definition aerial view over the area. We show that using a dataset of this size dramatically improves performance for key self-driving problems. Combined with the provided software kit, this collection forms the largest and most detailed dataset to date for the development of self-driving machine learning tasks, such as motion forecasting, motion planning and simulation. 
\end{abstract}



\section{Introduction}

The availability of large-scale datasets has been a large contributor to AI progress in the recent decade. In the field of self-driving vehicles (SDVs), several datasets, such as \cite{Geiger2013IJRR, 8953693, l5dataset}, enabled great progress within the development of perception systems \cite{DBLP:journals/corr/abs-1812-05784, liang2019multi, zhou2017voxelnet, DBLP:journals/corr/abs-1711-08488}. These allow an SDV to process LiDAR and camera sensors for understanding positions of other traffic participants including cars, pedestrians and cyclists around the vehicle.

Perception, however, is only the first step in the modern self-driving pipeline. Much work remains to be done around data-driven motion prediction of traffic participants, trajectory planning and simulation, before SDVs can become a reality. Datasets for developing these methods differ from those used for perception in that they require large amounts of behavioural observations and interactions. These are obtained by combining the output of perception systems with an understanding of the environment - in the form of a semantic map that contains priors over expected behaviour. Broad availability of datasets for these downstream tasks is much more limited though, and mostly available only to large-scale industrial efforts in the form of in-house collected data. This limits progress within the computer vision and robotics communities to advance modern machine learning systems for these important tasks.

In this work, we share the largest and most detailed dataset to date for training motion forecasting and planning solutions. We are motivated by the scenario of a self-driving fleet serving a single, high-demand route - rather than serving a broad area. We consider this to be a more feasible deployment strategy for ridesharing, since SDVs can be allocated to particular routes while human drivers serve the remaining traffic. This focus allows setting better bounds on required system performance and accident likelihood, both key factors for real-world self-driving deployment.

In summary, the released dataset consists of:
\begin{itemize}
    \item The largest dataset to date for motion prediction, containing 1,000 hours of traffic scenes that capture the motions of traffic participants around 20 self-driving vehicles, driving over 26,000 km along a suburban route.
    \item The most detailed high-definition (HD) semantic map of the area, counting over 15,000 human annotations including 8,500 lane segments.
    \item A high-resolution aerial image of the area, spanning 74 km$^2$ at a resolution of 6 cm per pixel, providing further spatial context about the environment.
    \item A Python software library L5Kit for accessing and visualising the dataset.
    \item Baseline machine learning solutions for the motion forecasting and motion planning task that demonstrate the impact of large-scale datasets.
\end{itemize}


\section{Related Work}

\begin{figure*}[b]
    \centering
    \includegraphics[width=\textwidth]{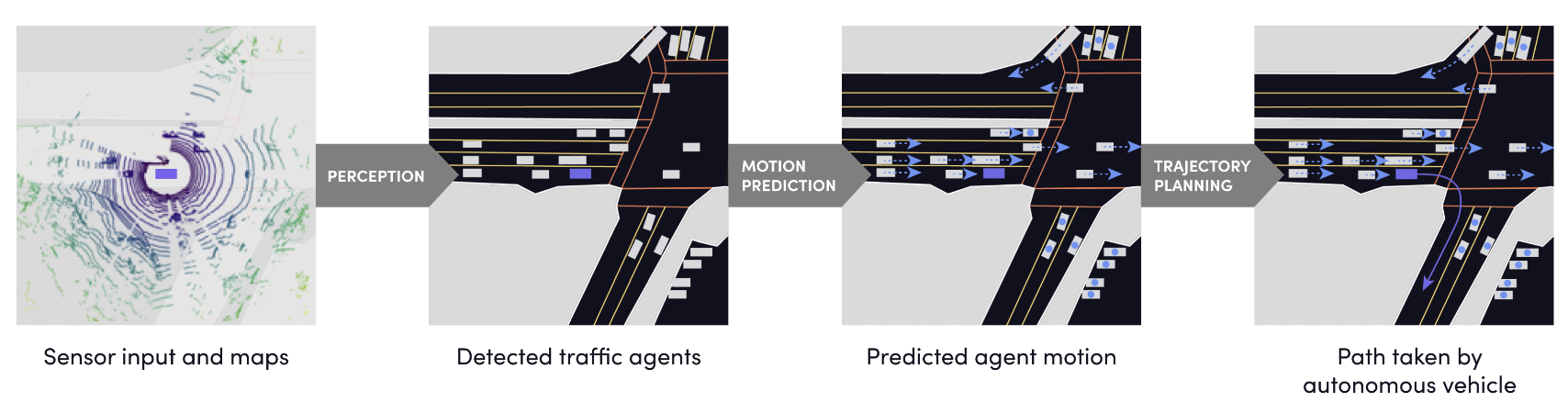}
    \caption{An example of a state-of-the-art self-driving pipeline. First, the raw LiDAR and camera data are processed to detect the positions of nearby objects around the vehicle. Then, their motion is predicted to allow the SDV to plan a safe collision-free trajectory. The released dataset enables the modelling of a motion prediction component.}
    \label{fig:pipeline}
\end{figure*}

\begin{table*}[t]
\small
    \centering
    \begin{tabular}{|l|l|l|l|l|l|}
    
    \hline
     \textbf{Name} & \textbf{Size} & \textbf{Scenes} & \textbf{Map} & \textbf{Annotations} & \textbf{Task}\\
    \hline
    \hline
        \multirow{2}{*}{KITTI \cite{Geiger2013IJRR}} & \multirow{2}{*}{6h} & \multirow{2}{*}{50} & \multirow{2}{*}{None} & 3D bounding & \multirow{2}{*}{Perception}\\
        & & & & boxes & \\
        \hline
        Oxford RobotCar \cite {RobotCarDatasetIJRR} & 71h & 100 & None & - & Perception\\
        \hline
        \multirow{2}{*}{Waymo Open Dataset \cite{sun2019scalability}} & \multirow{2}{*}{10h} & \multirow{2}{*}{1000} & \multirow{2}{*}{None} & 3D bounding & \multirow{2}{*}{Perception} \\
        & & & & boxes & \\
        \hline
        ApolloScape Scene & \multirow{2}{*}{2h} & \multirow{2}{*}{-} & \multirow{2}{*}{None} & 3D bounding & \multirow{2}{*}{Perception}\\
        Parsing \cite{Wang2019TheAO}  & & & & boxes & \\
        \hline
        Argoverse 3D Tracking  & \multirow{2}{*}{1h} & \multirow{2}{*}{113} & Lane center lines, & 3D bounding & \multirow{2}{*}{Perception} \\
        v1.1 \cite{8953693} & & & lane connectivity & boxes& \\
        \hline
       Lyft Perception Dataset  & \multirow{2}{*}{2.5h} & \multirow{2}{*}{366} & Rasterised & 3D bounding & \multirow{2}{*}{Perception}\\
        \cite{l5dataset} & & &  road geometry & boxes & \\
        \hline
        \multirow{2}{*}{nuScenes \cite{nuscenes2019}} & \multirow{2}{*}{6h} & \multirow{2}{*}{1000} & Rasterised & 3D bounding & Perception,\\
        & & & road geometry & boxes, trajectories & Prediction\\
        \hline
        ApolloScape Trajectory &  \multirow{2}{*}{2h} &  \multirow{2}{*}{103} &  \multirow{2}{*}{None} &  \multirow{2}{*}{Trajectories} &  \multirow{2}{*}{Prediction}\\
        \cite{Ma2019TrafficPredictTP}  & & & & & \\
        \hline
        Argoverse Forecasting &  \multirow{2}{*}{320h} &  \multirow{2}{*}{324k} & Lane center lines, &  \multirow{2}{*}{Trajectories} &  \multirow{2}{*}{Prediction} \\
        v1.1 \cite{8953693}  & & & lane connectivity & & \\
        \hline
    \hline
        \multirow{4}{*}{\textbf{Ours}} & \multirow{4}{*}{1,118h} & \multirow{4}{*}{170k} & Road geometry, & \multirow{4}{*}{Trajectories} & \\
        & & & aerial map, & & Prediction, \\
        & & & crosswalks, & & Planning \\
        & & & traffic lights state, ...  & & \\
    \hline
    \end{tabular}
    \caption{A comparison of various self-driving datasets available today. Our dataset surpasses all others in terms of size, as well as level of detail of the semantic map (see Section \ref{sec:ds}).}
    \label{tab:datasets}
\end{table*}

In this section we review related existing datasets for training SDV systems from the viewpoint of a classical state-of-the-art self-driving stack as summarised in Figure \ref{fig:pipeline}. In this stack the raw sensor input is first processed by a perception system to estimate positions of nearby vehicles, pedestrians, cyclists and other traffic participants. Next, the future motion and intent of these actors is estimated (also called motion prediction or forecasting) and used for planning SDV trajectory.
In Table \ref{tab:datasets} we summarise the current leading datasets for training machine learning solutions for different components of this stack, focusing mainly on the perception and prediction components.

\textbf{Perception datasets}
The perception task is usually framed as the supervised task of estimating 3D positions of nearby objects around the SDV. Deep learning approaches are now state-of-the-art for most subproblems relevant for autonomous driving, such as 3D object detection and semantic segmentation \cite{DBLP:journals/corr/abs-1711-08488, zhou2017voxelnet, liang2019multi, DBLP:journals/corr/abs-1812-05784}.

Among the datasets for training these systems the KITTI dataset \cite{Geiger2013IJRR} is the most common benchmarking dataset for many computer vision and autonomous driving related tasks. It contains around 6 hours of driving data, recorded from front-facing stereo cameras, LiDAR and GPS/IMU sensors. 3D bounding box annotations are available, including class annotations such as cars, trucks and pedestrians.
The Waymo Open Dataset \cite{sun2019scalability} and nuScenes \cite{nuscenes2019} are of similar size and structure, providing 3D bounding box labels based on fused sensory inputs. 
The Oxford RobotCar dataset \cite{RobotCarDatasetIJRR} also allows application for visual tasks, but focus more lies on localisation and mapping, rather than object detection.

Our dataset's main target is not to train perception systems. Instead, it is a product of an already trained perception system used to process large quantities of new data for motion prediction.

\textbf{Prediction datasets}
The prediction task we focus on in this paper builds on top of perception by trying to predict positions of detected objects a few seconds into the future. In order to obtain good results, one needs significantly more detailed information about the environment including, for example, semantic maps that encode possible driving behaviour to reason about future behaviours.

Deep learning solutions leveraging birds-eye-view (BEV) representations of scenes \cite{7780479, gupta2018social, Lee2017DESIREDF, Cui2019MultimodalTP, chai2019multipath, hong2019rules} or graph neural networks \cite{casas2019spatiallyaware, gao2020vectornet} have established themselves as the leading solutions for this task. Representative large-scale datasets for training these systems are, however, rare. The above mentioned solutions were developed almost exclusively by industrial labs leveraging internal proprietary datasets.

The most relevant existing open dataset is the Argoverse Forecasting dataset \cite{8953693} providing 300 hours of perception data and a lightweight HD semantic map encoding lane center positions. Our dataset differs in three substantial ways: 1) Instead of focusing on a wide city area we provide 1000 hours of data along a single route. This is motivated by the assumption that, particularly in ride-hailing applications, the first deployments of self-driving fleets are more likely to occur along few high-demand routes. This makes it possible to set bounds for requirements and quantify accident risk more precisely. 2) We are contributing higher-quality scene data by providing the full perception output including bounding boxes and class probabilities instead of pure centroids. In addition, our semantic map is more detailed: it counts more than 15,000 human annotations instead of only lane centers. 3) We further provide a high-resolution aerial image of the area. This is motivated by the fact that much of the information encoded in the semantic map is implicitly  accessible in the aerial form. Providing this map can, therefore, unlock the development of semantic-map free solutions.

\textbf{Planning datasets}
Planning and executing an SDV trajectory is the final component in the autonomous driving stack. It is also the component having received the least attention from the community and exhibited the least progress when counting ML solutions.
One of the reasons is that it is difficult to model and evaluate this task as an ML problem trained from real data. Closed-loop evaluation of a driving policy requires collecting new data that are not present in the dataset.

As a result, a majority of research work instead focuses on open-loop ego-motion forecasting \cite{Zeng2019EndToEndIN, philion2020learning}. However, it is known that such methods significantly underperform in closed-loop evaluation due to the distribution shift \cite{ross2010efficient}. Consequentially, most of state-of-the-art solutions used in industry resort to traditional trajectory optimisation systems based on hand-crafted cost functions instead of machine learning.

In our work we follow the recent approach \cite{bansal2018chauffeurnet} leveraging imitation learning and perturbations. Authors, however, demonstrated this approach only on a proprietary dataset with no available open equivalent. As part of our work we provide an ML planning baseline inspired by \cite{bansal2018chauffeurnet} and show it can be effectively trained and evaluated using our dataset, yielding the first such open evaluation. We show that, similarly to motion forecasting, the performance of ML planning significantly improves with the amount of training data - showing much promise for ML planning solutions to the future together with the need of datasets to train them.


\begin{figure*}[b]
    \centering
    \includegraphics[width=140mm]{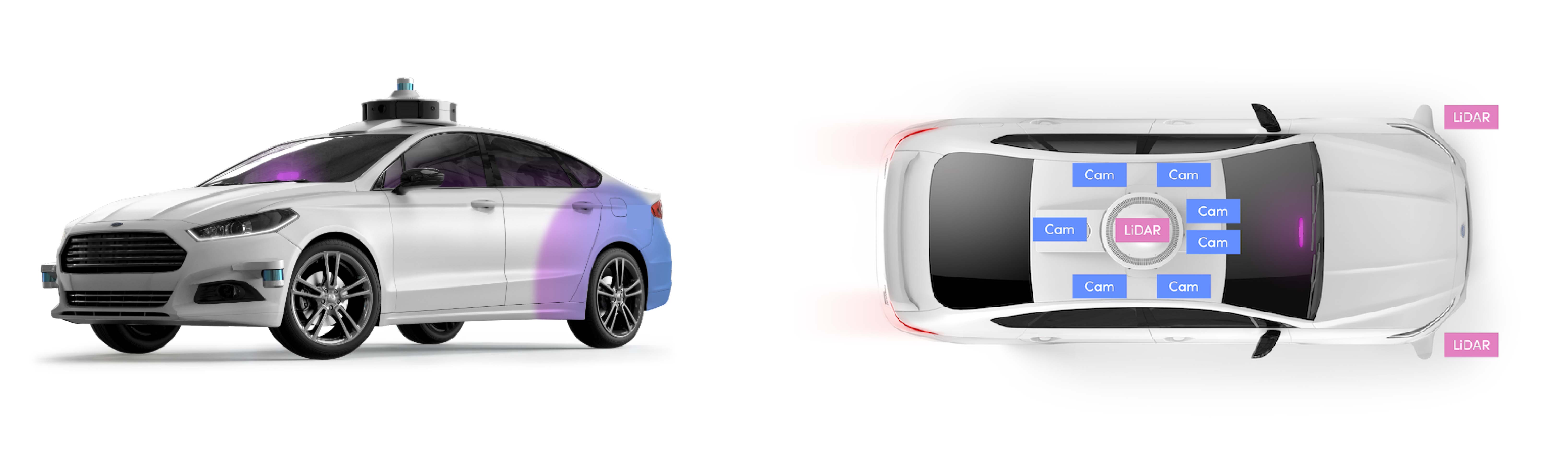}
    \caption{The self-driving vehicle configuration used to collect the data. Raw data from LiDARs and cameras were processed by a perception system to generate the dataset, capturing the poses and motion of nearby vehicles.}
    \label{fig:hardware}
\end{figure*}


\begin{figure*}[t]
    \centering
    \begin{tabular}{cccccc}
    \includegraphics[width=19mm]{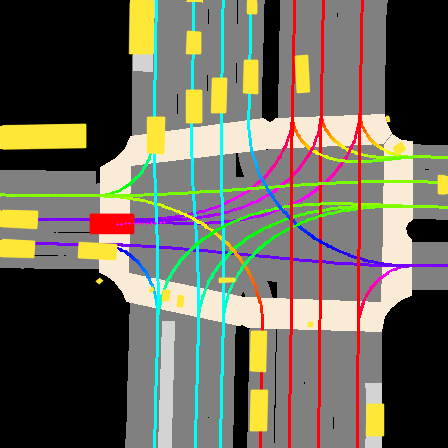} &
    \includegraphics[width=19mm]{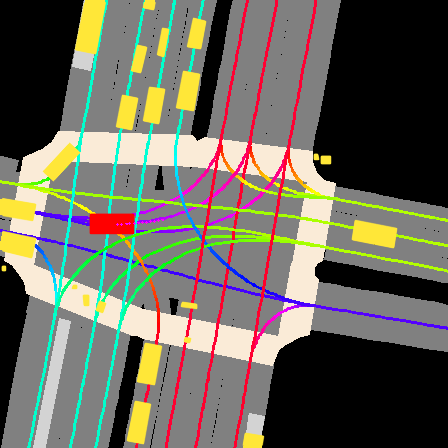} &
    \includegraphics[width=19mm]{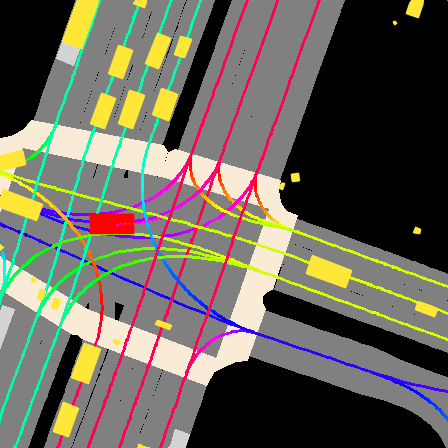} &
    \includegraphics[width=19mm]{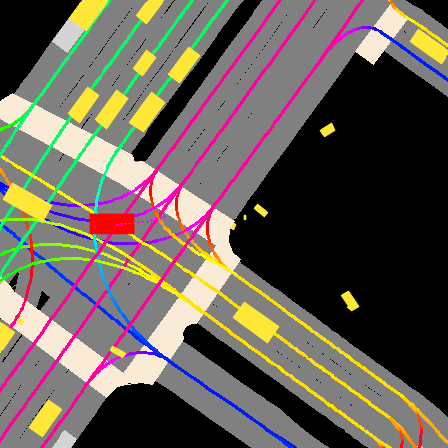} &
    \includegraphics[width=19mm]{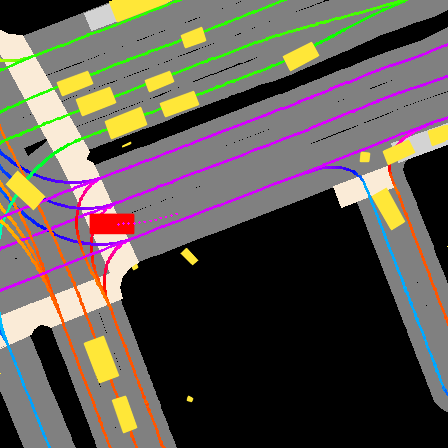} &
    \includegraphics[width=19mm]{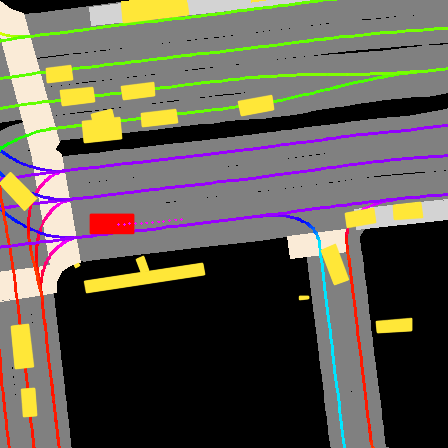} \vspace{3mm}\\
    \includegraphics[width=19mm]{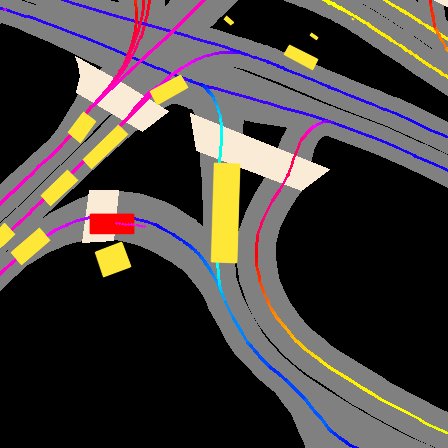} &
    \includegraphics[width=19mm]{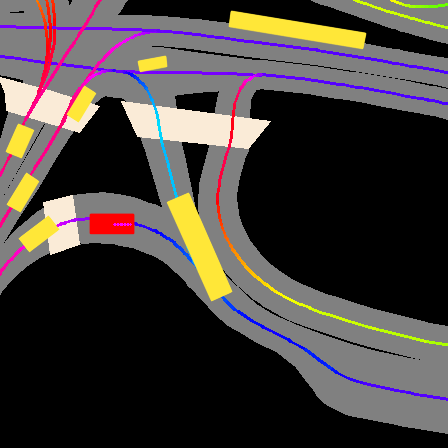} &
    \includegraphics[width=19mm]{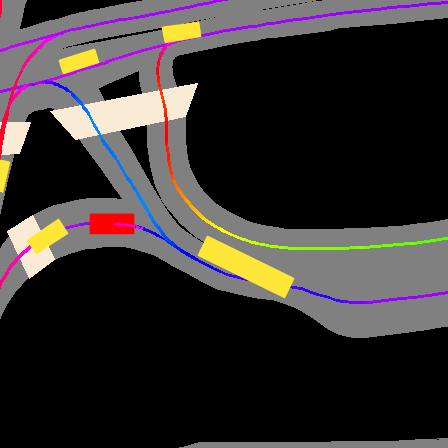} &
    \includegraphics[width=19mm]{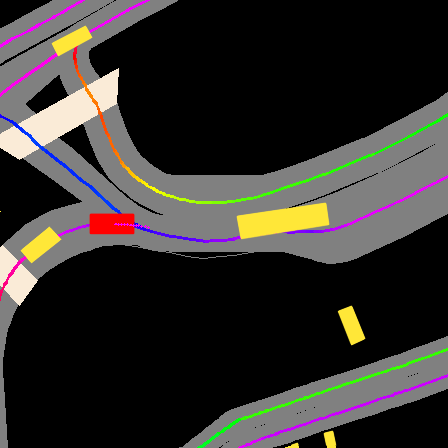} &
    \includegraphics[width=19mm]{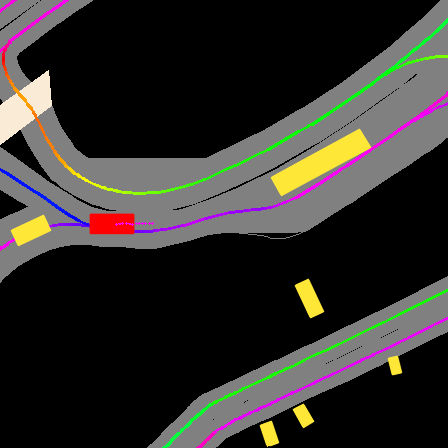} &
    \includegraphics[width=19mm]{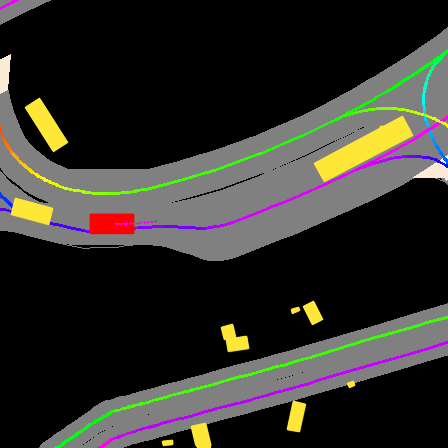} \vspace{3mm} \\
    \includegraphics[width=19mm]{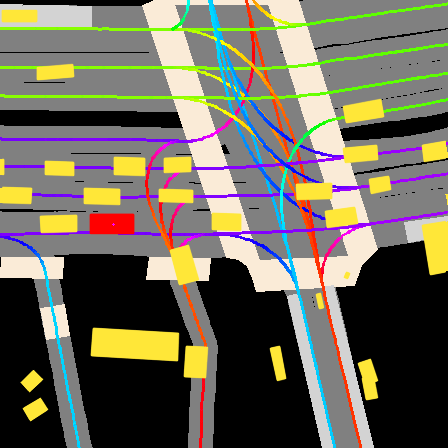} &
    \includegraphics[width=19mm]{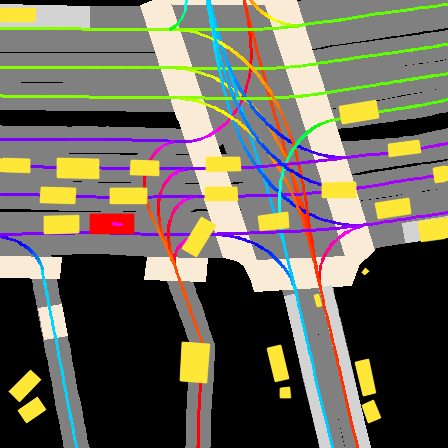} &
    \includegraphics[width=19mm]{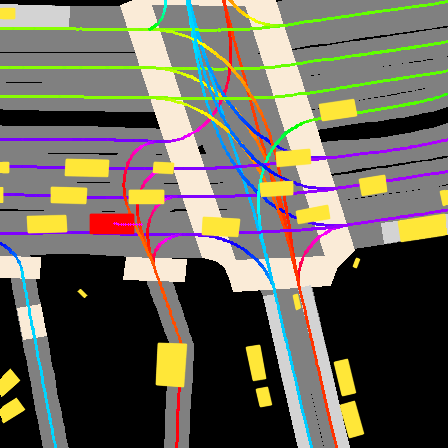} &
    \includegraphics[width=19mm]{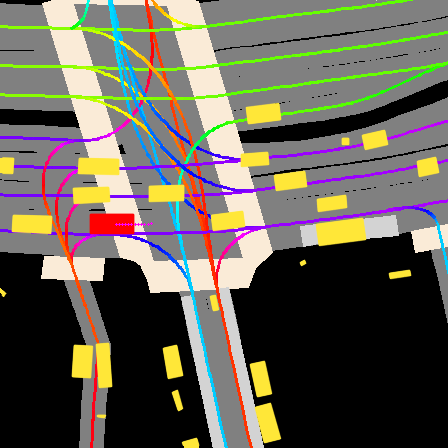} &
    \includegraphics[width=19mm]{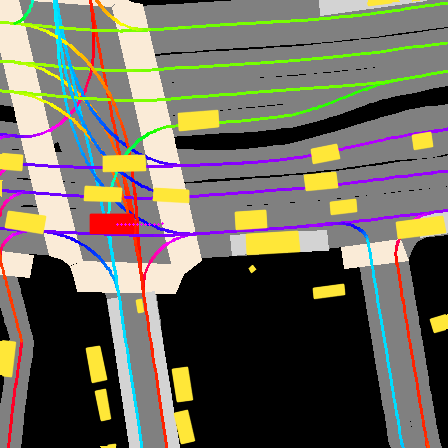} &
    \includegraphics[width=19mm]{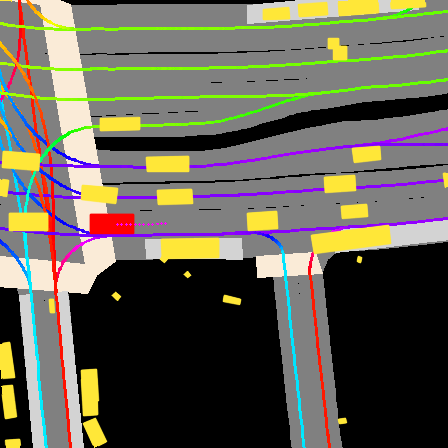} \vspace{3mm} \\
    \includegraphics[width=19mm]{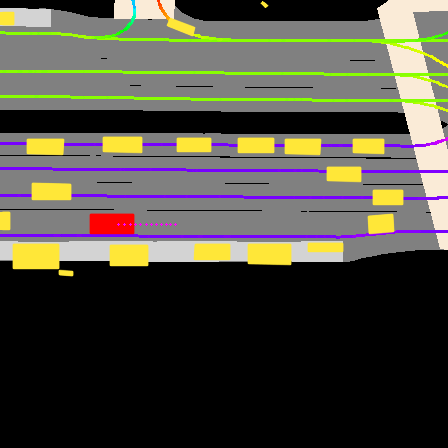} &
    \includegraphics[width=19mm]{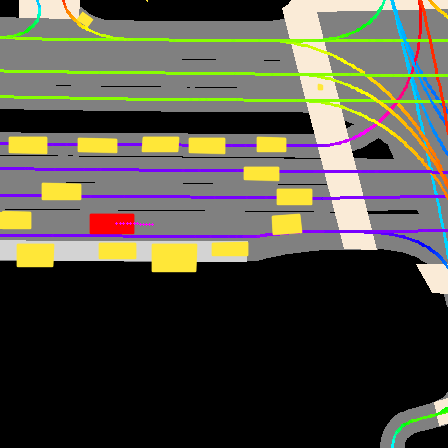} &
    \includegraphics[width=19mm]{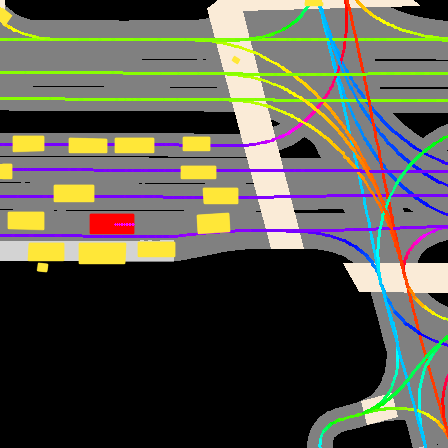} &
    \includegraphics[width=19mm]{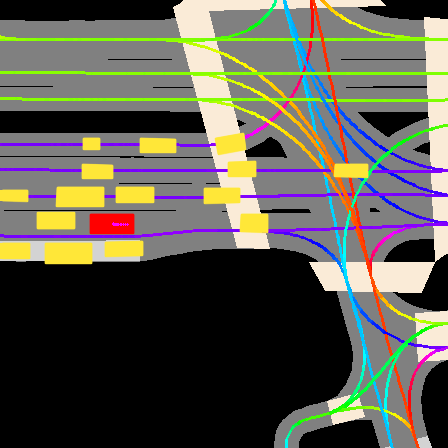} &
    \includegraphics[width=19mm]{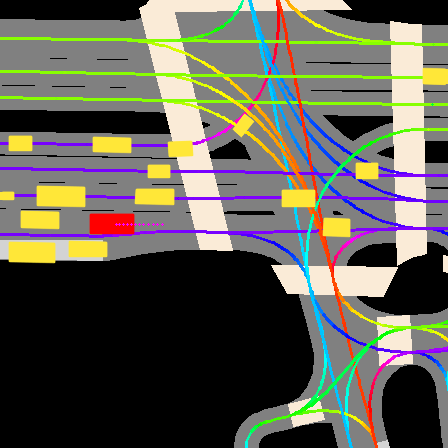} &
    \includegraphics[width=19mm]{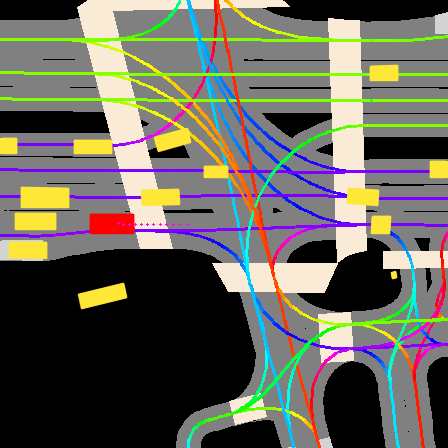} \\
    \end{tabular}
    \caption{Examples from the scenes in the dataset, projected over a BEV of the rasterised semantic map. The self-driving vehicle is shown in red, other traffic participants in yellow, and lane colours denotes driving direction. The dataset contains 170,000 such sequences, each 25 seconds long with sensor data at 10Hz.}
    \label{fig:data}
\end{figure*}


\section{Dataset}
\label{sec:ds}

Here we outline the details of the released dataset, including the process that was used to construct it. An overview of different dataset statistics can be found in Table \ref{tab:stat}.

The dataset has three components:
\begin{enumerate}
    \item 170,000 scenes, each 25 seconds long, capturing the movement of the self-driving vehicle, traffic participants around it and traffic lights state.
    \item A high-definition semantic map capturing the road rules, lane geometry and other traffic elements.
    \item A high-resolution aerial picture of the area that can be used to further aid in prediction.
\end{enumerate}
\begin{table*}
\centering
\begin{tabular}{|l|l|}
\hline
\textbf{Statistic}            & \textbf{Value}           \\
\hline
\hline
\# self driving vehicles used & 20\\
Total data set size & 1,118 hours / 26,344 km / 162k scenes \\
Training set size          & 928 hours / 21,849 km / 134k scenes           \\
Validation set size            & 78 hours / 1,840 km / 11k scenes \\
Test set size     & 112 hours / 2,656 km / 16k scenes             \\
\hline
Scene length                  & 25 seconds               \\
Total \# of traffic participant observations      & 3,187,838,149            \\
Average \# of detections per frame      & 79                 \\
Labels & Car: 92.47\% / Pedestrian: 5.91\% / Cyclist: 1.62\%       \\
\hline
Semantic map & 15,242 annotations / 8,505 lane segments \\
Aerial map & 74 km$^2$ at 6 cm per pixel\\
\hline
\end{tabular}
\caption{Statistics of the released dataset.}
\label{tab:stat}
\end{table*}

\subsection{Scenes}
The dataset consists of 170,000 scenes, each 25 seconds long, totalling over 1,118 hours of logs. Example scenes are shown in Figure \ref{fig:data}. All logs were collected by a fleet of self-driving vehicles driving along a fixed route. The sensors for perception include 7 cameras, 3 LiDARs, and 5 radars (see Figure \ref{fig:hardware}). The sensors are positioned as follows: one LiDAR is on the roof of the vehicle, and two LiDARs on the front bumper. The roof LiDAR has 64 channels and spins at 10 Hz, while the bumper LiDARs have 40 channels. All seven cameras are mounted on the roof and together have a 360$^{\circ}$ horizontal field of view. Four radars are also mounted on the roof, and one radar is placed on the forward-facing front bumper. 

The dataset was collected between October 2019 and March 2020. It was captured during daytime, between 8 AM and 4 PM. For each scene we detected the visible traffic participants, including vehicles, pedestrians, and cyclists. Each traffic participant is internally represented by a 2.5D cuboid, velocity, acceleration, yaw, yaw rate, and a class label. These traffic participants are detected using our in-house perception system similar to \cite{DBLP:journals/corr/abs-1711-08488, zhou2017voxelnet, liang2019multi, DBLP:journals/corr/abs-1812-05784} and trained using a dataset similar to \cite{l5dataset} along the same route. It fuses data across multiple modalities to produce a 360$^{\circ}$ view of the world surrounding the SDV. Table \ref{tab:stat} outlines some more statistics for the dataset.

We split the dataset into train, validation and test set using a 83--7--10\% ratio, where a particular SDV only contributes to a single set. We encode the dataset in the form of $n$-dimensional compressed zarr arrays. The zarr format\footnote{  \url{https://zarr.readthedocs.io/}} was chosen to represent individual scenes. It allows for fast random access to different portions of the dataset while minimising the memory footprint, which allows efficient distributed training on the cloud.

\subsection{High-definition semantic map}

The HD semantic map that we provide encodes information about the road itself as well as various traffic elements along the route totalling 15,242 labelled elements, including 8,505 lane segments. This map was created by human curators who annotated the underlying localisation map, which in turn was created using a simultaneous localisation and mapping (SLAM) system. Given the use of SLAM, the position of the SDV is always known with centimetre-grade accuracy. Thus, the information in the semantic map can be used both for planning driving behaviour and for anticipating the future movements of other traffic participants.

\begin{figure*}[t]
    \begin{minipage}{9cm}
    \begin{tabular}{|l|l|}
    \hline
    \textbf{Property} & \textbf{Values}\\
    \hline
    \hline
    Lane boundaries & sequence of $(x,y,z)$ coordinates \\
    Lane connectivity & possible lane transitions \\
    Driving directions & one way, two way \\
    Road class & primary, secondary, tertiary, ... \\
    Road paintings & solid, dashed, colour \\
    Speed limits & mph \\
    Lane restrictions & bus only, bike only, turn only, ... \\
    Crosswalks & position \\
    Traffic lights & position, lane association\\
    Traffic signs & stop, turn, yield, parking, ...\\
    Restrictions & keep clear zones, no parking, ...\\
    Speed bumps & position \\
    \hline
    \end{tabular}
    \end{minipage}
    \begin{minipage}{4.6cm}
    \includegraphics[width=\textwidth]{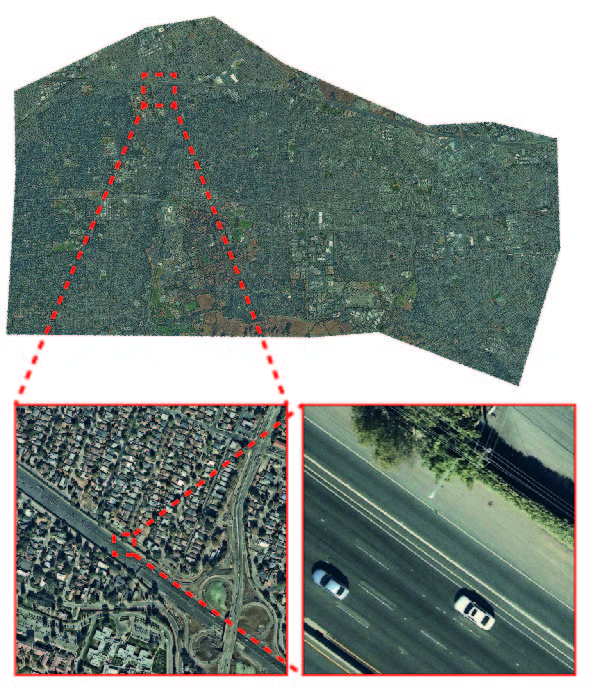}
    \end{minipage}
    \caption{Elements of the provided HD semantic map (left) and overhead aerial map surrounding the route (right). We provide 15,242 human annotations including 8,505 individual lane segments. The aerial map covers 74 km$^2$ at a resolution of 6 cm per pixel.}
    \label{fig:map}
\end{figure*}

The semantic map is given in the form of a protocol buffer\footnote{\url{https://developers.google.com/protocol-buffers}}. We provide precise road geometry through the encoding of the lane segments, their connectivity and other properties (as summarised in Figure \ref{fig:map}).

\subsection{Aerial map}
The aerial map captures the area of Palo Alto surrounding the route at a resolution of 6 cm per pixel. It enables the use of spatial information to aid with motion prediction. Figure \ref{fig:map} shows the map coverage and the level of detail. The covered area of 74 km$^2$ is provided as 181 GeoTIFF tiles of size $10560 \times 10560$ pixels, each spanning approximately $640 \times 640$ meters.

\section{Development tools}

\begin{figure*}[!h]
	\centering
	\begin{minipage}[t]{6.5cm}
	\centering
        \includegraphics[width=65mm]{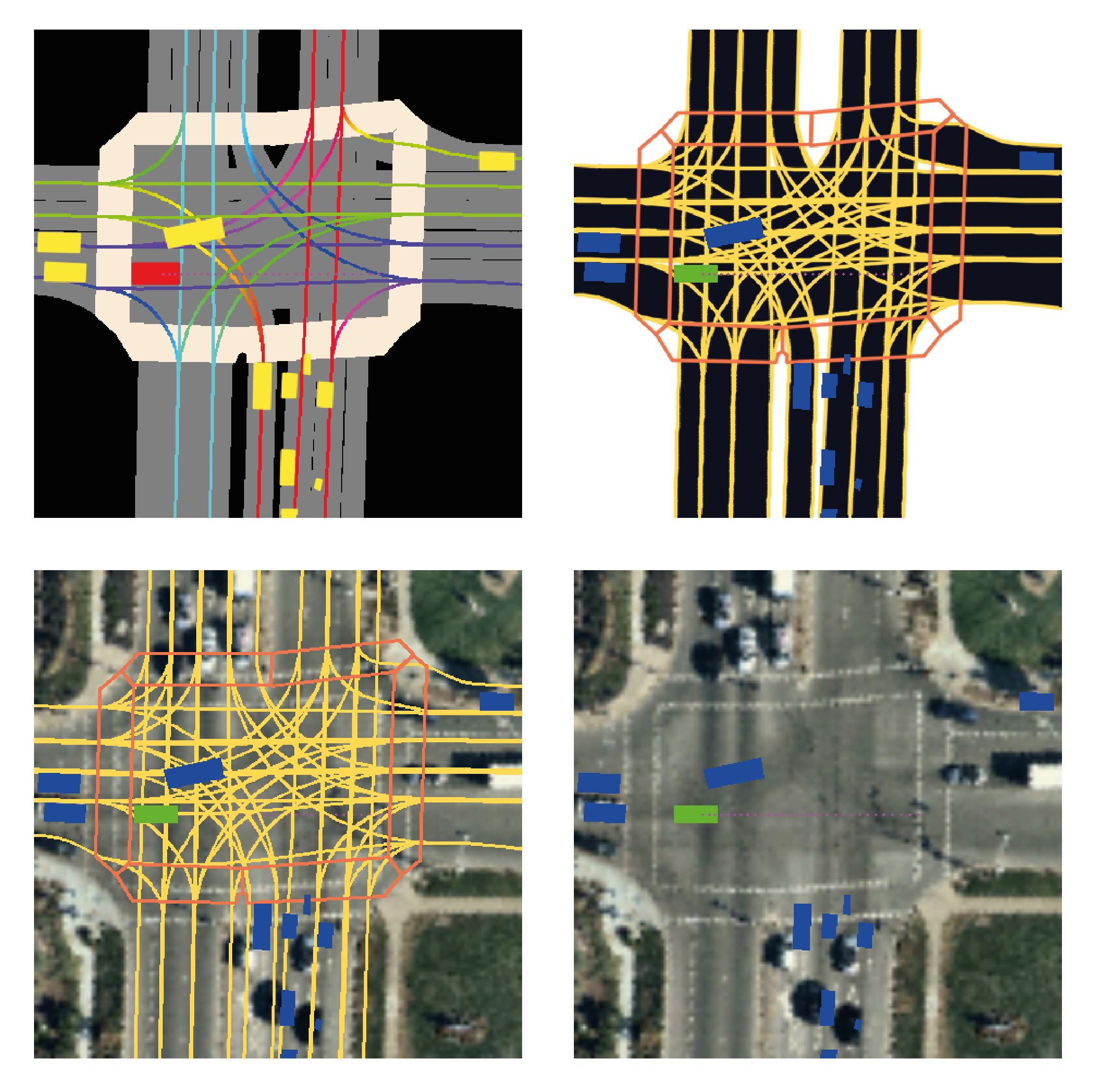}
        \caption{Examples of different BEV scene rasterisations that can be made using the associated software development kit. These can be used, for example, as input to convolutional neural network architectures.}
        \label{fig:rasterisation}
	\end{minipage}
	\hspace{3mm}
	\begin{minipage}[t]{6.7cm}
        \centering
        \includegraphics[width=65mm]{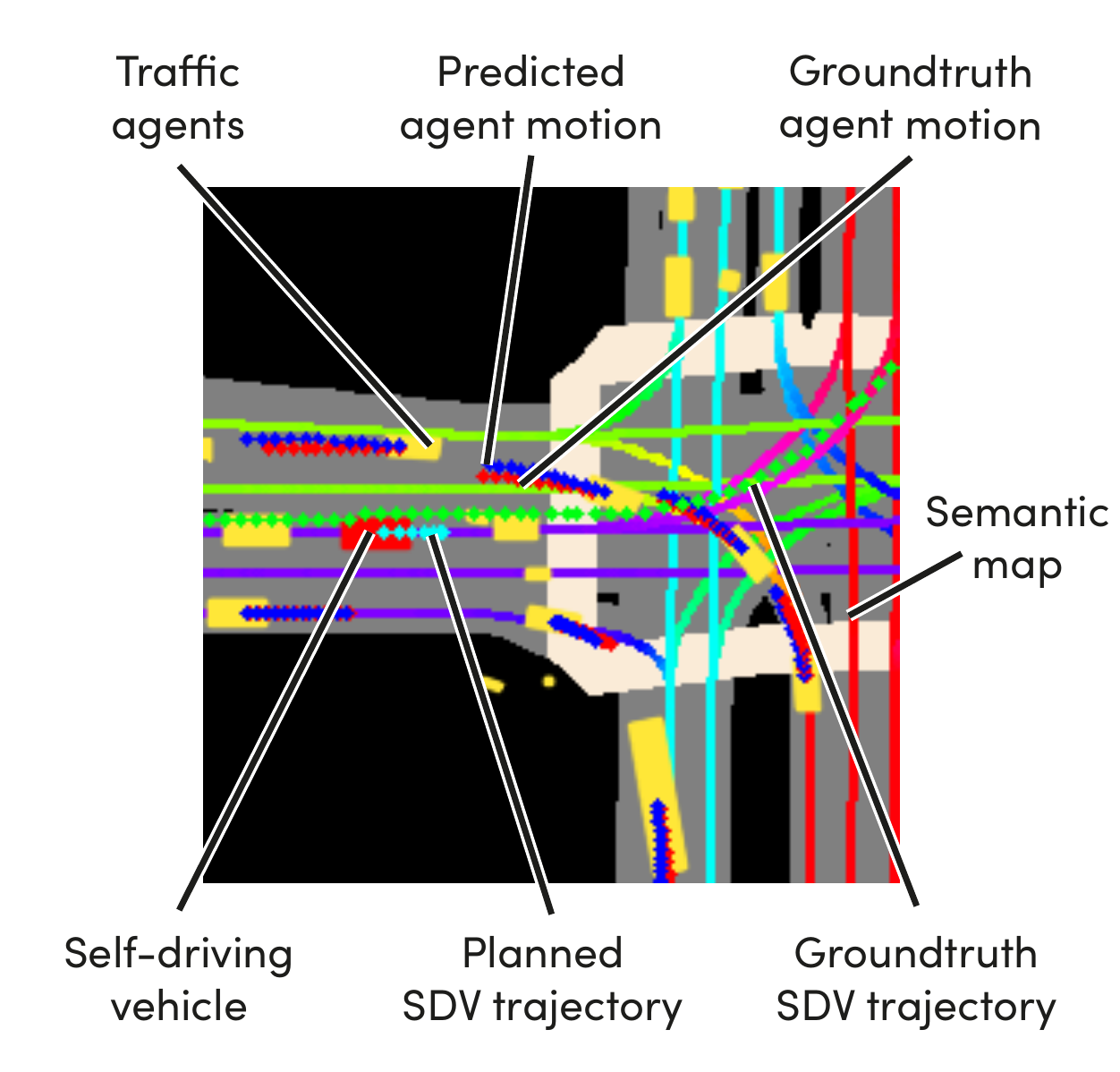}
        \caption{Example output of the agent motion forecasting and SDV motion planning baseline, supplied as part of the software development kit.}
        \label{fig:prediction}
	\end{minipage}
\end{figure*}

In combination with the dataset, we are releasing a Python toolkit called L5Kit\footnote{\url{https://github.com/lyft/l5kit}}. It provides access to data loading and visualisation functionality and implementation for two baseline tasks of motion forecasting and SDV motion planning.

\textbf{Multi-threaded data loading and sampling} 
We provide an API that can sample scenes and load the data efficiently. Scenes can be sampled from multiple points of view: for planning of the SDV motion path, we can center the scene around the SDV. For predicting the motions of other traffic participants, we provide functionality to recenter the scene around those traffic participants.

\textbf{Customisable BEV scene rasterisation} 

We provide several functions to visualise and rasterise a sampled scene. Our visualisation package can draw additional information, such as the future trajectory, onto an RGB image and save files as images, GIFs or full scene videos. 

We support several different rasterisation modes for creating a meaningful representation for the underlying image. Figure \ref{fig:rasterisation} shows example images generated by the different modes and created from either the semantic map (upper right image) or the aerial map (lower right), or a combination of both (lower left). Such images can then be used as input to a conventional machine learning pipeline akin to \cite{bansal2018chauffeurnet, Cui2019MultimodalTP}.

\begin{table*}[b]
\centering
\begin{tabular}{ll|lllllll}
\multicolumn{2}{c}{\textbf{Configuration}} & \multicolumn{7}{c}{\textbf{Displacement error [m]}} \\
 Training size & History length & @0.5s & @1s & @2s & @3s & @4s & @5s & ADE\\
\hline
1\% & 0 sec & 0.44 & 0.84 & 1.62 & 2.41 & 3.26 & 4.22 & \textbf{2.47}\\
10\% & 0 sec & 0.36 & 0.69 & 1.29 & 1.91 & 2.57 & 3.30 & \textbf{1.95}\\
100\% & 0 sec & 0.31 & 0.59 & 1.10 & 1.61 & 2.14 & 2.74 & \textbf{1.64}\\
\hline
1\% & 1 sec & 0.16 & 0.30 & 0.59 & 0.97 & 1.47 & 2.08 & \textbf{1.08}\\
10\% & 1 sec & 0.15 & 0.27 & 0.54 & 0.87 & 1.30 & 1.84 & \textbf{0.96}\\
100\% & 1 sec & 0.13 & 0.23 & 0.44 & 0.70 & 1.03 & 1.46 & \textbf{0.77}
\end{tabular}
\caption{Performance of the motion forecasting baseline in open-loop evaluation. The performance continues increasing with the size of the training set, both for models using history and not using it. We list the displacement error for different prediction horizons.}
\label{tab:results_prediction}
\end{table*}

\textbf{Motion forecasting baseline}
In motion forecasting, the task is to predict the expected future (x,y)-positions over a $T=5$-second-horizon for different traffic participants in the scene given their current (and sometimes also historical) positions. We use a ResNet-50 backbone \cite{7780459} with $L_2$ loss that was trained on $224 \times 224$ pixel BEV rasters centered around different vehicles of interest. We also provide a history of the vehicles' movements over the past few seconds by simply stacking BEV rasters. This allows the network to implicitly compute an agent's current velocity and heading. Figure \ref{fig:prediction} displays typical predictions after training each vehicle on this architecture for 38,000 iterations with a batch size of 64.

Table \ref{tab:results_prediction} summarises the displacement error (the $L_2$-norm between the predicted point and the true position at horizon $T$) for a network trained on the semantic map at various prediction horizons for different history lengths , as well as the displacement error averaged over all timesteps (ADE). To evaluate the impact of the size of the dataset, we train models using different sample sizes from the training set. As seen, adding history improves the prediction performance as the network gains knowledge about vehicle speed and acceleration. Moreover, the performance keeps increasing with the size of the training set.

\textbf{Motion planning baseline}
As a related experiment, we employ our dataset for the SDV planning task. The task here is to predict and also execute a trajectory for the SDV. This allows the actual SDV pose to be different in future timesteps than recorded in the log. As outlined in \cite{ross2010efficient} using motion forecasting is not a suitable solution for this problem as it suffers from accumulation of errors due to broken i.i.d assumption. Our implementation is based on \cite{bansal2018chauffeurnet} of augmenting motion forecasting model with synthetic perturbations to mitigate this problem. The network is trained to accept BEV rasters of size $224 \times 224$ pixels  (centered around the SDV this time) to predict future (x,y)-positions over a 5-second-horizon. We use the same architecture and loss as in the motion forecasting task, and train for a similar number of iterations.

\begin{figure*}
    \centering
    \includegraphics[width=\textwidth]{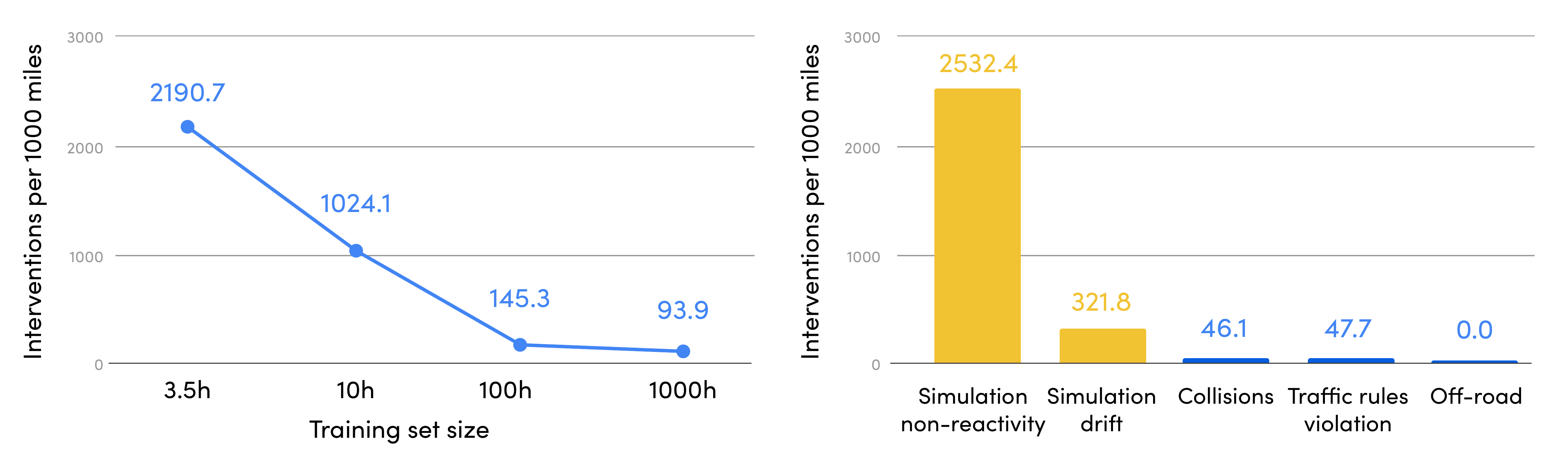}
    \caption{Performance of the ML planning task in a closed-loop evaluation. The SDV is free to take actions and diverge from the recorded behaviour while other traffic participants follow the recorded log. The left plot shows the performance of the SDV increasing dramatically with the amount of training data. Right is the qualitative performance of a model trained on 1000h.}
    \label{fig:results_planning}
\end{figure*}

Testing of the system constitutes simulating SDV behaviour in 25 sec. long scenes from the test dataset. In each episode each traffic participant follows logged behaviour. The SDV is, however, controlled by the network and is free to take any action and diverge from its original behaviour. Results are summarised in Figure \ref{fig:results_planning}. This simulation is not perfect as it is non-reactive, but still gives valuable insights into SDV's actual performance. We count only errors caused by the SDV's actions (collisions, traffic rules violations, off-road events) and not limitations of blind log-replay simulation (e.g. other cars colliding into the SDV due to non-reactivity, diverging too far from the log position rendering perception ineffective), which dominate the total errors. Similarly to motion forecasting the performance keeps increasing with more data. This suggests that the performance is not saturated and both tasks can benefit from even larger datasets in the future.



\section{Conclusion}

The dataset introduced in this paper is the largest and most detailed dataset available for training prediction and planning solutions. It is three times larger and significantly more descriptive than the current best alternatives. We show that this difference yields a meaningful increase in performance for both the motion forecasting and motion planning task. This is in-line with the intuition, that datasets are key ingredients in unlocking and making large-scale machine learning systems work well. These datasets, however, are not available for everyone as they often come from proprietary industrial efforts.

We believe that publishing this dataset marks an important next step towards the democratisation within the development of self-driving applications. This, in turn, can result in faster progress towards a fully autonomous future. At the same time, we observe the performance of the motion forecasting and motion plannning tasks still keeps increasing with the size of the training data. This suggests, that even larger datasets counting tens of thousands or even millions of hours can be desirable in the future, together with algorithms that can take advantage of them.



\acknowledgments{This work was done thanks to many members of the Lyft Level 5 team. Specifically, we would like to thank Emil Praun, Christy Robertson, Oliver Scheel, Stefanie Speichert, Liam Kelly, Chih Hu, Usman Muhammad, Lei Zhang, Dmytro Korduban, Jason Zhao, Hugo Grimmett and Luca Del Pero.}


\bibliography{main}  

\end{document}